\documentclass[manuscript,screen]{acmart}
\usepackage{graphicx}
\usepackage{subcaption}
\usepackage[linesnumbered,ruled,vlined,noend]{algorithm2e}
\usepackage{multirow}
\usepackage{enumerate}
\usepackage{adjustbox}
\graphicspath{{images/}}

\AtBeginDocument{%
  \providecommand\BibTeX{{%
    \normalfont B\kern-0.5em{\scshape i\kern-0.25em b}\kern-0.8em\TeX}}}

\setcopyright{acmlicensed}
\acmJournal{TOMM}
\acmYear{2022} \acmVolume{1} \acmNumber{1} \acmArticle{1} \acmMonth{1} \acmPrice{15.00}\acmDOI{10.1145/3531013}
\acmConference[]{}{}{}
\acmBooktitle{}
\acmPrice{15.00}
\acmISBN{978-1-4503-XXXX-X/18/06}

\begin{document}

\title{Optimizing Performance of Federated Person Re-identification: Benchmarking and Analysis}

\author{Weiming Zhuang}
\email{weiming001@e.ntu.edu.sg}
\orcid{0000-0001-8243-7772}
\affiliation{
  \institution{S-Lab, Nanyang Technological University}
  \streetaddress{50 Nanyang Ave}
  \city{Singapore}
  \country{Singapore}
  \postcode{639798}
}
\additionalaffiliation{
    \institution{SenseTime Research}
    \country{China}
}

\author{Xin Gan}
\email{ganx0005@e.ntu.edu.sg}
\affiliation{
  \institution{Nanyang Technological University}
  \streetaddress{50 Nanyang Ave}
  \city{Singapore}
  \country{Singapore}
  \postcode{639798}
}

\author{Yonggang Wen}
\email{ygwen@ntu.edu.sg}
\affiliation{
  \institution{Nanyang Technological University}
  \streetaddress{50 Nanyang Ave}
  \city{Singapore}
  \country{Singapore}
  \postcode{639798}
}

\author{Shuai Zhang}
\affiliation{
    \institution{SenseTime Research}
    \country{China}
}
\email{zhangshuai@sensetime.com}

\renewcommand{\authors}{Weiming Zhuang, Yonggang Wen, Xuesen Zhang, Xin Gan, and Shuai Zhang}

\renewcommand{\shortauthors}{W. Zhuang et al.}

\begin{abstract}

The increasingly stringent data privacy regulations limit the development of person re-identification (ReID) because person ReID training requires centralizing an enormous amount of data that contains sensitive personal information. To address this problem, we introduce federated person re-identification (\textit{FedReID}) --- implementing federated learning, an emerging distributed training method, to person ReID. FedReID preserves data privacy by aggregating model updates, instead of raw data, from clients to a central server. Furthermore, we optimize the performance of FedReID under statistical heterogeneity via benchmark analysis. We first construct a benchmark with an enhanced algorithm, two architectures, and nine person ReID datasets with large variances to simulate the real-world statistical heterogeneity. The benchmark results present insights and bottlenecks of FedReID under statistical heterogeneity, including challenges in convergence and poor performance on datasets with large volumes. Based on these insights, we propose three optimization approaches: (1) We adopt \textit{knowledge distillation} to facilitate the convergence of FedReID by better transferring knowledge from clients to the server; (2) We introduce \textit{client clustering} to improve the performance of large datasets by aggregating clients with similar data distributions; (3) We propose \textit{cosine distance weight} to elevate performance by dynamically updating the weights for aggregation depending on how well models are trained in clients. Extensive experiments demonstrate that these approaches achieve satisfying convergence with much better performance on all datasets. We believe that FedReID will shed light on implementing and optimizing federated learning on more computer vision applications.

\end{abstract}



\maketitle

\section{Introduction}

Person re-identification (ReID) aims to match the same person who appeared in disjoint camera views. It has received considerable attention because of its wide applications in business and public security, such as customer trajectory analysis and criminal investigation \cite{leng2019survey}. Person ReID has achieved outstanding performances \cite{sun2018beyond, chen2019abd, ye2021deep-survey}, attributing to the advances of deep neural networks (DNN) \cite{krizhevsky2012alexnet, resnet}.

However, the increasing concerns of data privacy protection limit the development of person ReID \cite{gdpr}. DNN-based approaches are data-hungry, which relies on centralizing a sizable amount of data to achieve high performance \cite{zheng2016person-reid-survey}. Training images of person ReID contains sensitive personal information, which could reveal the identity and location of individuals. Centralizing these images impose potential privacy leakage risks. Hence, it is crucial to navigate the development of person ReID under the premise of privacy protection.

Federated learning (FL), an emerging distributed training technique, has empowered many applications with privacy-preserving mechanisms \cite{kairouz2019advances}, such as healthcare applications \cite{sheller2018brain-tumor2, chen2020fedhealth} and consumer products \cite{niu2020recommendation-alibaba, muhammad2020fedfast}. FL preserves data privacy by training models collectively with decentralized clients. These clients, instead of transferring raw data, only transfer training updates to a central server. It reduces privacy leakage risks as raw data are kept locally. Despite the advantages of FL, implementing FL to person ReID and optimizing its performance are largely overlooked; such implementation possibility is only mentioned in \cite{Hao2018EdgeAIBench}, but that study does not present dataset or benchmark results.

In this work, we propose Federated Person Re-identification (\textit{FedReID}), a new person ReID training paradigm to enable multimedia researchers to train models with privacy guaranteed. Besides privacy protection, FedReID possesses other advantages: reducing communication overhead of uploading plenty amount of data \cite{McMahanMRHA17}; adapting models in clients to local scenes; obtaining a holistic model that generalizes in diverse scenarios. A usage example of FedReID is video surveillance across communities or districts, where multiple entities collaborate to learn a generalized model without revealing their private video surveillance data. 

However, implementing FL to person ReID is not trivial --- statistical heterogeneity is a major challenge of FedReID in real-world scenarios \cite{Li2020FedChallenges}: 1) data is in \textit{non-identical and independent distribution (non-IID)} \cite{zhao2018non-iid} because data collected from different cameras could have significant discrepancies in resolution, illumination, and angles; 2) data volume is \textit{unbalanced} with varied pedestrian flow in different locations. Although some studies illustrate that non-IID harms the training convergence and model performance in tasks like image classification \cite{zhao2018non-iid}, the impact of statistical heterogeneity on FedReID is not explored before.

This work aims to optimize FedReID under statistical heterogeneity via benchmark analysis. We start by constructing a new benchmark, FedReIDBench, with nine representative ReID datasets and a specially designed algorithm for FedReID (Sec. \ref{sec:benchmark}). In the benchmark, a server coordinates nine clients (each containing a dataset) to conduct training on their local data and aggregates training updates iteratively. We then conduct benchmark analysis (Sec. \ref{sec:benchmark-analysis}), revealing that statistical heterogeneity leads to performance degradation and difficulty in convergence. We end by proposing three performance optimization methods: client clustering (Sec. \ref{sec:clustering}) and dynamic weight adjustment (Sec. \ref{sec:weight-adjustment}) to elevate performance; knowledge distillation (Sec. \ref{sec:kd}) to facilitate convergence. 
Specifically, \textit{client clustering} groups clients with similar data distribution and aggregate training updates within each group. \textit{Knowledge distillation} uses a public dataset to transfer knowledge from clients to the server more effectively. Besides, \textit{weight adjustment} dynamically updates the weights of clients' training updates in server aggregation. 
Extensive experiments demonstrate the effectiveness of the benchmark and the significance of the optimization approaches. We believe that FedReID will shed light on implementing and optimizing federated learning to more computer vision applications.

In summary, we make the following contributions: 

\begin{itemize}
    \item We construct a new benchmark for Federated Person Re-identification (FedReID), simulating real-world scenarios of statistical heterogeneity with 9 representative person ReID datasets.
    \item We provide useful insights and investigate potential bottlenecks of FedReID by analyzing the benchmark results.
    \item We propose three performance optimization methods: knowledge distillation to facilitate convergence; client clustering and dynamic weight adjustment to elevate performance.
    \item We extensively evaluate these optimization methods to demonstrate their effectiveness.
\end{itemize}


\section{Related Works}\
\label{sec:related-work}


\subsection{Person Re-Identification}

The objective of person ReID is to retrieve the identity of interest from disjoint camera views. It is an important computer vision task that is widely applied in public security, such as video surveillance \cite{zheng2016person-reid-survey}. The advances of the deep neural network have greatly improved the performance of person ReID by learning better feature representations, compared to traditional hand-crafted feature operations \cite{2014deepReID, Liu-reid-multi-scale, local-cnn-reid-acmmm, multi-gra-acmmm}. Over the years of development, the community has constructed many person ReID datasets \cite{Wei2017Msmt, Zheng2015Market1501, zheng2017dukemtmc-reid, li2012cuhk01, prid2011}. These datasets are collected from various locations with different camera views. The majority of person ReID studies focus on extracting better feature representations by improving the architecture of deep neural networks \cite{leng2019survey, ye2021deep-survey}. They rely on the assumption that data, collected from different cameras in various locations, can be centralized to a central server. However, centralizing plenty of images of individuals raises potential privacy leakage risks. Different from previous approaches, we proposed FedReID --- a new training paradigm for ReID to learn ReID models from decentralized data. FedReID mitigates potential privacy leakage issues as data is not transferred to a central server.


\subsection{Federated Learning}

Federated learning (FL) is an emerging distributed training technique that trains models with decentralized clients coordinated by a central server \cite{kairouz2019advances}.

\textbf{Benchmarks} To facilitate the development of FL, researchers have published several benchmarks and datasets: LEAF \cite{Leaf} is the first benchmark for FL research, containing federated datasets for image classification and natural language processing tasks; Streets datasets \cite{luo2019realworld} is a real-world image dataset collected from street cameras for object detection; OARF \cite{hu2020oarf} is a benchmark aims to facilitate a wide range of FL applications, such as trend prediction, recommendation, and sentiment analysis. However, different from these tasks, person ReID is a retrieval task that no existing benchmark contains related datasets. In this work, we construct a new FL benchmark that simulates real-world scenarios of FedReID.


\textbf{Algorithm} The most known algorithm for FL is Federated Averaging (FedAvg) \cite{McMahanMRHA17}. It defines an iterative training process that clients send trained local models to a server and the server sends back the aggregated global model to clients. Benchmarks mentioned above adopt FedAvg as the standard algorithm. However, FedAvg requires all clients to have identical models. It is not suitable for FedReID because clients could have varied classifiers. Therefore, we propose an enhanced algorithm, Federated Partial Averaging (FedPav).

\textbf{Statistical Heterogeneity} Statistical heterogeneity --- non-IID and unbalanced data --- is a major challenge of FL \cite{Li2020FedChallenges, kairouz2019advances}. In traditional distributed training \cite{large-scale-dnn-NIPS2012, sunpeng}, data in multiple nodes of cloud clusters are IID. Data in multiple FL clients, however, could be heterogeneous. To address this challenge, some studies focus on optimizing training in clients \cite{fedprox, acar2020feddyn, li2021moon, karimireddy2020scaffold, zhang2021fedufo}; while the recent work \cite{zhang2021fedufo} requires extra communication by sharing features among clients. On the other hand, some studies optimize the aggregation process in the server \cite{fedma,wang2020fednova, zhuang2021fedu, zhuang2022fedssl}. Besides, several studies share voluntary or public data between the server and clients \cite{zhao2018non-iid, xin2019meta}. These methods are validated on small datasets \cite{krizhevsky2009cifar10,cohen2017emnist,Leaf}, and thus may not be directly applicable to the challenging scenario of FedReID. In this work, we introduce three optimization methods targeting the statistical heterogeneity of FedReID via in-depth benchmark analysis.


This work is an extension of our previous conference version \cite{zhuang2020fedreid}. The main improvements are as follows: 1) We introduce a new performance optimization method --- client clustering; 2) We integrate client clustering with the previously proposed weight adjustment method, achieving the best performance; 3) We conduct more performance evaluations for comparison with the benchmark results and the proposed optimization methods; 4) We provide more comprehensive descriptions for the proposed optimization methods. Despite that another work \cite{wu2021fedreid} also studies FedReID after our conference work \cite{zhuang2020fedreid}, it focuses more on adapting to unseen domains, while we aim to address the statistical heterogeneity revealed from our benchmark analysis.




\begin{table}[t]
\caption{The statistics of nine datasets in our proposed benchmark, FedReIDBench. These datasets have large variances in data volume, decreasing from top to bottom.}
\begin{tabular}{lccccccccc}
\toprule
\multicolumn{1}{l}{\multirow{3}{*}{Datasets}} &
  \multicolumn{1}{c}{\multirow{3}{*}{\# Cameras}} &
  \multicolumn{2}{c}{Train} &
  \multicolumn{1}{c}{} &
  \multicolumn{5}{c}{Test} \\ \cline{3-4}  \cline{6-10}  
\multicolumn{1}{c}{} &
  \multicolumn{1}{c}{} &
  \multirow{2}{*}{\# IDs} &
  \multirow{2}{*}{\# Images} &
  \multirow{2}{*}{} &
  \multicolumn{2}{c}{Query} &
  \multirow{2}{*}{} &
  \multicolumn{2}{c}{Gallery} \\ \cline{6-7} \cline{9-10} 
\multicolumn{1}{c}{} &
  \multicolumn{1}{c}{} & & & & \# IDs & \# Images & & \# IDs & \# Images \\ 
\midrule
MSMT17 \cite{Wei2017Msmt} & 15 & 1,041 & 32,621 & & 3,060 & 11,659 & & 3,060 & 82,161 \\ 
DukeMTMC-reID \cite{zheng2017dukemtmc-reid} & 8 & 702 & 16,522 & & 702 & 2,228 & & 1,110 & 17,611 \\
Market-1501 \cite{Zheng2015Market1501} & 6 & 751 & 12,936 & & 750 & 3,368 & & 751 & 19,732 \\ 
CUHK03-NP \cite{Li2014CUHK03} & 2 & 767 & 7,365 & & 700 & 1,400 & & 700 & 5,332 \\ 
PRID2011 \cite{prid2011} & 2 & 285 & 3,744 & & 100 & 100 & & 649 & 649 \\ 
CUHK01 \cite{li2012cuhk01} & 2 & 485 & 1,940 & & 486 & 972 & & 486 & 972 \\ 
VIPeR \cite{Gray2008ViewpointIP} & 2 & 316 & 632 & & 316 & 316 & & 316 & 316 \\ 
3DPeS \cite{3dpes} & 2 & 93 & 450 & & 86 & 246 & & 100 & 316 \\ 
iLIDS-VID \cite{iLIDS-VID} & 2 & 59 & 248 & & 60 & 98 & & 60 & 130 \\ 
\bottomrule
\end{tabular}
\label{tab:dataset}
\end{table}

\section{Federated Person ReID Benchmark}
\label{sec:benchmark}

This section introduces a new federated learning benchmark for person ReID, FedReIDBench. This benchmark comprises nine representative datasets, two possible implementation architectures, one enhanced algorithm, and several performance evaluation metrics.

\subsection{Datasets}

We construct the benchmark dataset with nine representative person ReID datasets as shown in Table \ref{tab:dataset}. It contains in total 224,064 images of 17,991 identities. These datasets are collected at multiple locations (or countries) and published by different organizations at different times. They not only vary in the number of images, identities, and camera views, but also differ in image resolution, illumination, and scenes.

The variances in these datasets simulate the statistical heterogeneity in real-world scenarios: the disparity of data volumes represents the unbalanced data problem; the domain discrepancies among datasets represent the non-IID problem. Unlike centralized training where data is IID, statistical heterogeneity makes training even more challenging. 

\subsection{Architectures}

\begin{figure}
  \begin{minipage}{0.3\linewidth}
      \begin{subfigure}[t]{0.9\textwidth}
        \centering
        \includegraphics[width=1\textwidth]{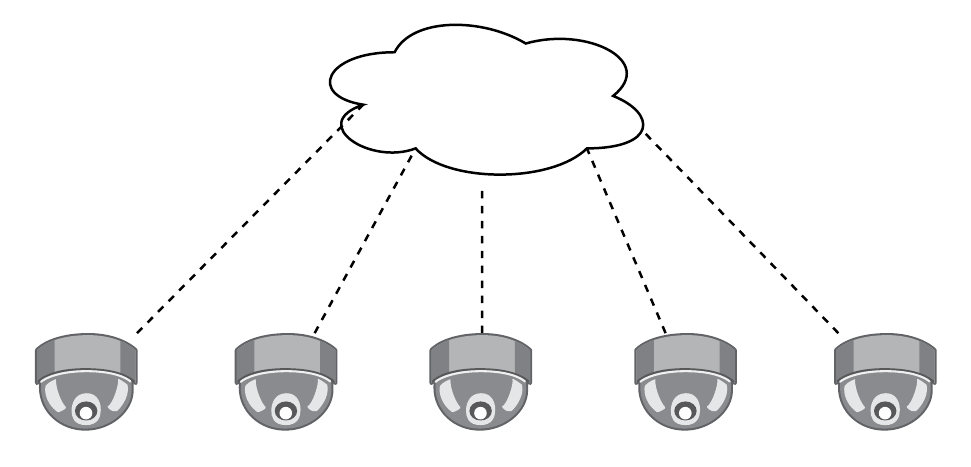}
        \caption{Edge-cloud Architecture}
        \label{fig:edge-cloud}
    \end{subfigure}
    \begin{subfigure}[t]{0.9\textwidth}
        \centering
        \includegraphics[width=1\textwidth]{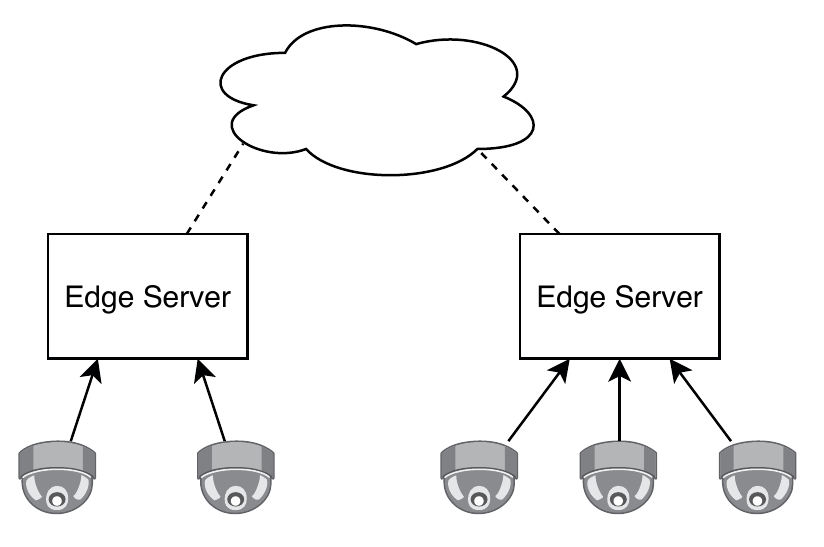}
        \caption{Device-edge-cloud Architecture}
        \label{fig:device-edge-cloud}
    \end{subfigure}
  \end{minipage}%
  \hfill
  \begin{minipage}{0.7\linewidth}
      \begin{subfigure}[t]{1\textwidth}
        \centering
        \includegraphics[width=1\textwidth]{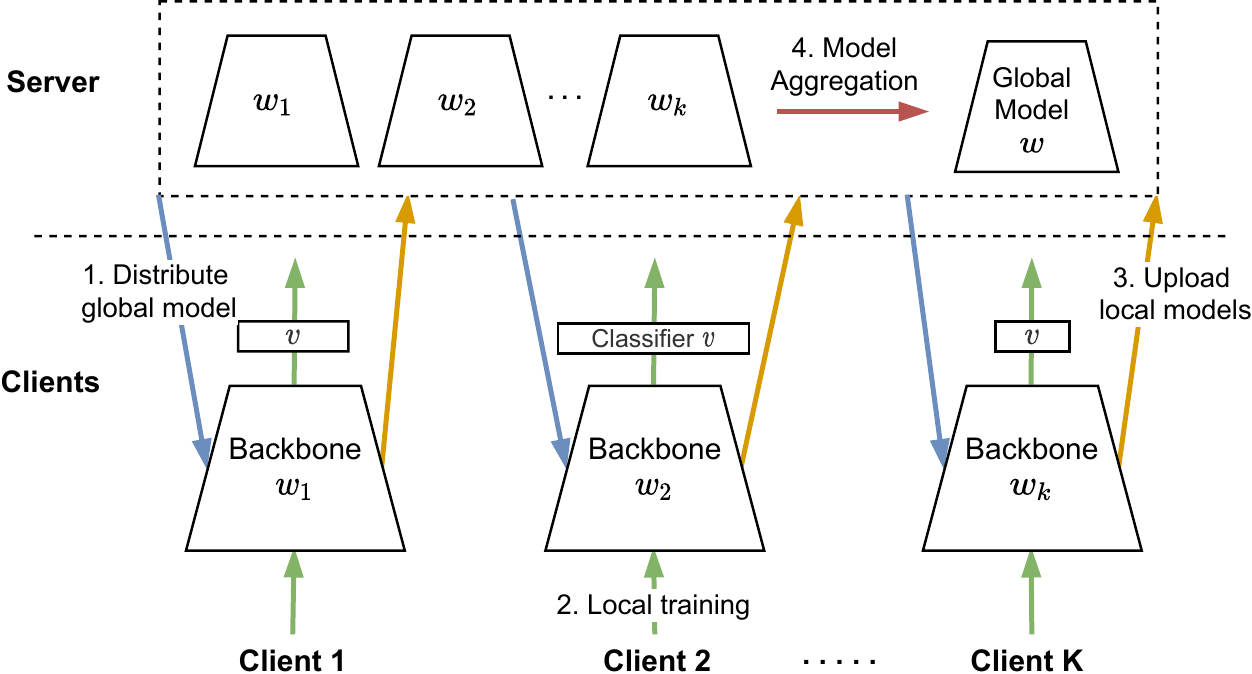}
        \caption{Federated Partial Averaging}
        \label{fig:fedpav}
    \end{subfigure}
  \end{minipage}
  \caption{Illustrations of FedReID architectures and benchmark algorithm Federated Partial Averaging (FedPav). (a) Edge-cloud architecture and (b) device-edge-cloud architecture are two possible architectures for FedReID. (c) FedPav algorithm is suitable for both architectures, regarding cameras as clients for (a) and edge servers as clients for (b). Each training round of FedPav has four steps: 1) A central server distributes a global model to selected clients; 2) Clients conduct training with their local data; 3) Clients upload trained models; 4) The server obtains a new global model by aggregating these models.}
\end{figure}

Figure \ref{fig:edge-cloud} and Figure \ref{fig:device-edge-cloud} illustrate two architectures for possible implementation scenarios of FedReID: edge-cloud architecture and device-edge-cloud architecture. In both architectures, the cloud represents the central server connecting to multiple edges.

\textbf{Edge-cloud Architecture}: In this architecture, cameras are the edges that directly connect with the server to conduct federated learning. The server coordinates these cameras to train models with locally collected images. This architecture significantly reduces privacy leakage risks as the data always stays at the edges. However, deployment of this architecture requires cameras to have enough computation power and storage capability. A real-world application of this architecture would be video surveillance for a community with multiple cameras on different streets. 

\textbf{Device-edge-cloud Architecture}: This is a three-layer hierarchical architecture. Edge servers are in the middle layer. On the one hand, they construct local training datasets by gathering images from multiple camera views, which is similar to how datasets in the benchmark are collected. On the other hand, edge servers collaboratively perform federated learning with their local datasets under the coordination of the server. A good illustration of this architecture would be multiple communities collaborate to learn person ReID models, where each community has an edge server collecting data from multiple cameras.

\SetKwInput{KwInput}{Input}                
\SetKwInput{KwOutput}{Output}              
\SetKwFunction{FnClient}{}
\SetKwFunction{FnServer}{}
\SetKwComment{Comment}{$\triangleright$\ }{}

\begin{algorithm}[]
    \caption{Federated Partial Averaging (FedPav)}
    \label{algo:fedpav}
    \SetAlgoLined
    \KwInput{Total data volume $n$, client $k$'s data volume $n_k$, local epoch $E$, batch size $B$}
    \KwOutput{Global model $w^T$, local models of each client $w^T_k$}
    
    \SetKwProg{Fn}{Server}{:}{}
    \Fn{\FnServer}{
        initialize $w^0$ \;
        \For{each round t = 0 to T-1}{
            $S_t \leftarrow$ (random set of K clients selected from N clients) \Comment*[r]{Client selection}
            \For{each client $k \in S_t$ in parallel}{
                $w^{t+1}_k \leftarrow $ \textbf{Client}($w^t$, $k$, $t$) \Comment*[r]{Distribution}
            }
            $w^{t+1} \leftarrow \sum_{k \in C_t} \frac{n_k}{n} w^{t+1}_k$
            \Comment*[r]{Server aggregation}
        }
        \KwRet best global model $w^{T}$, best client $k$'s local model $w^{T}_k$\;
    }
    
    \SetKwProg{Fn}{Client}{:}{\KwRet}
    \Fn{\FnClient{w, k, t}}{
        $v \leftarrow$ (initialize classifier $v$ if $t == 0$, otherwise retrieve it from local store)\;
        \For{each epoch e = 0 to E-1}{
            \For{each batch $b \in$ batches of local data in size $B$}{
                $(w, v) \leftarrow (w, v) - \eta \triangledown \mathcal{L}((w, v); b)$ \Comment*[r]{$(w, v)$ concatenation of two model structures}
            }
        }
        store $v$\;
        \KwRet $w$ \Comment*[r]{Upload local model}
    }
\end{algorithm}

\subsection{Algorithm}

The standard federated learning algorithm Federated Averaging \cite{McMahanMRHA17} is not suitable for FedReID because it requires identical model structures in all clients. The model structure of the benchmark is ID-discriminative embedding (IDE) \cite{zheng2016person-reid-survey}, which is a common baseline for DNN-based person ReID. This model structure consists of a backbone and a classifier: the backbone is ResNet-50 \cite{resnet} in our FedReIDBench; the classifier is a linear layer whose dimension depends on the number of identities of clients. Since the number of identities could vary among clients, their classifiers could differ in clients. Hence, we adopt the enhanced algorithm for FedReID: Federated Partial Averaging \cite{zhuang2020fedreid}.

Federated Partial Averaging (FedPav) allows models in clients to be only partially identical. For FedReID, FedPav enables clients to use the same backbone but different identity classifiers for federated learning, as shown in Figure \ref{fig:fedpav}. The training process is similar to FedAvg except that clients only transfer the identical part of models to the central server for aggregation.

Algorithm \ref{algo:fedpav} summarizes FedPav. We aim to obtain a holistic global model and personalized local models for clients at the end of the training. Each training round $t$ of FedPav contains four steps: 1) \textit{Distribution}: the central server chooses a fraction ($K$ out of $N$) of clients for current round of training and distributes the global model $w^t$ to these clients; 2) \textit{Local training}: each client $k$ initializes the backbone $w_k^{t}$ using the global model parameters and train the model with local dataset for $E$ local epochs with $B$ batch size; 3) \textit{Upload}: each client $k$ uploads the trained backbone $w_k^{t+1}$ to the server; 4) \textit{Aggregation}: the server generates a new global model $w^{t+1}$ by aggregating updates from clients with weighted average.
The training stops after iterating these four steps for $T$ rounds.
After training, we use the global model $w$ to evaluate convergence and generalization, and use local models $w_k$ to evaluate how well models adapt to local scenarios.


\subsection{Performance Evaluation Metrics}

We evaluate FedReID in two aspects: accuracy and communication cost. 

\textbf{Accuracy} Cumulative Matching Characteristics (CMC) curve and mean Average Precision (mAP) \cite{zheng2016person-reid-survey} are standard person ReID evaluation metrics. Given an image as a query, person ReID matches it in a gallery of images based on similarity. CMC measures the probability that the query identity is in the top-$k$ most similar matched gallery images. We consider $k=\{1, 5, 10\}$ in the benchmark, representing the rank-1 accuracy, rank-5 accuracy, and rank-10 accuracy. Besides, we also report the mAP of all queries.

\textbf{Communication Cost} Since federated learning requires iterative communication between a server and multiple clients, we also consider the communication costs. The total communication cost is $T \times 2 \times M$, where $T$ is the number of communication rounds and $M$ is the transmission message size (model size). $2 \times M$ is the communication cost of each round, considering both uploading and downloading from clients. 


\subsection{Reference Implementation}
\label{sec:referenced-implementation}

To facilitate ease-of-use and reproducibility, we open-source referenced implementation in GitHub \footnote{https://github.com/cap-ntu/FedReID}. It includes data preprocessing, proposed algorithm, and optimization methods. We plan to integrate it to EasyFL \cite{zhuang2022easyfl} in the future. Besdies, we provide the experimental settings as follows:

\textbf{Learning Rate} The initialized learning rates were different for the identity classifier and the backbone: 0.05 for the identity classifier; 0.005 for the backbone. The learning rate schedulers of both are the same with step size 40 and gamma 0.1. Besides, the learning rate for the server fine-tuning in knowledge distillation is 0.0005.

\textbf{Optimizer} We use Stochastic Gradient Decent (SGD) as the optimizer. The optimizer is set with weight decay 5e-4 and momentum 0.9.

\textbf{Federated Learning Settings} The default settings of federated learning algorithms are as follows: batch size $B = 32$, local epoch $E = 1$, and total training rounds $T = 300$.

\section{Benchmark Analysis}
\label{sec:benchmark-analysis}

In this section, we present the results of extensive experiments on the benchmark. We investigate the performance of two architectures, the impact of different federated settings, and the impact of statistical heterogeneity.




We initialize the backbone with ResNet-50\cite{resnet} parameters pre-trained on ImageNet \cite{imagenet_cvpr09}. For hyperparameters, we use batch size $B = 32$ and local epoch $E = 1$ to train $T = 300$ communication rounds by default. 

%

\subsection{Edge-cloud Architecture}


In the edge-cloud architecture, each camera is a client. Since each person ReID dataset contains data from several camera views, we simulate FedReID in this architecture by assigning data of the same camera view to one client. As a dataset is divided into several clients by camera views, we term it \textit{federated-by-camera} scenario. 

To understand FedReID performance in the federated-by-camera scenario, we compare it with two other settings: 1) \textit{Federated-by-identity} scenario: we divide one dataset into partitions for multiple clients, where each client includes one partition that contains an equal number of identities. The number of clients equals the number of camera views. 2) \textit{Centralized training}: training with data merged from multiple cameras, which can be considered as the upper bound. For example, Market-1501 dataset \cite{Zheng2015Market1501} contains six camera views with 751 identities. In the federated-by-identity scenario, we divide it into six clients, where each client includes 125 non-overlapping identities. The centralized training means training with the Market-1501 dataset.



Table \ref{tab:edge-cloud-arch} presents the comparisons of global models of different settings on two datasets: CUHK03-NP dataset \cite{Li2014CUHK03} and Market-1501 dataset \cite{Zheng2015Market1501}. Compared with the federated-by-identity scenario or centralized training, the federated-by-camera scenario performs much worse. It indicates that learning from only one camera view is infeasible to obtain a generalized model in person ReID, where the evaluation is based on images from multiple camera views. Hence, even though industrial cameras have enough computation and storage capacity to support edge-cloud architecture, the device-edge-cloud architecture could be more adequate for FedReID because each client learns cross-camera knowledge. All the other experiments in the paper are conducted based on the device-edge-cloud architecture.


\begin{table}[t]
\caption{Performance comparison of federated-by-camera scenario, federated-by-identity scenario, and centralized training on CUHK03-NP and Market-1501 datasets. The federated-by-camera scenario achieves the worst performance, indicating that edge-cloud architecture could be inadequate for FedReID.}
\begin{tabular}{cclcccccc}
\toprule
\multicolumn{1}{c}{Dataset} & \# Clients & Settings & Rank-1 & Rank-5 & Rank-10 & mAP \\ 
\midrule

\multirow{3}{*}{CUHK03-NP} & \multirow{3}{*}{2} & Federated-by-camera & 11.21 & 19.14 & 25.71 & 11.11 \\
\multicolumn{1}{l}{} & & Federated-by-identity & \textbf{51.71} & \textbf{69.50} & \textbf{76.79} & \textbf{47.39} \\
\multicolumn{1}{l}{} & & Centralized Training & 49.29 & 68.86 & 76.57 & 44.52 \\ 

\midrule

\multirow{3}{*}{Market-1501} & \multirow{3}{*}{6} & Federated-by-camera & 61.13 & 74.88 & 80.55 & 36.57 \\
\multicolumn{1}{l}{} & & Federated-by-identity & 85.69 & 93.44 & 95.81 & 66.36 \\
\multicolumn{1}{l}{} & & Centralized Training & \textbf{88.93} & \textbf{95.34} & \textbf{96.88} & \textbf{72.62} \\
\bottomrule
\end{tabular}
\label{tab:edge-cloud-arch}
\end{table}

\subsection{Device-Edge-Cloud Architecture}
\label{sec:federated-by-datasets}

In the device-edge-cloud architecture, edge servers collect data from multiple cameras and conduct FedReID with a central server. Since each of the benchmark datasets consists of data from multiple camera views, we simulate this scenario with nine clients --- each client contains one unique dataset of the benchmark datasets. In all experiments, we choose nine clients to participate in training.

Under this architecture, We consider two types of models produced from FedReID training: 1) \textit{Local model}: the specialized models trained after $E$ local epochs in clients before uploading to the server in each training round. 2) \textit{Global model}: the generalized model obtained in the server by aggregating models uploaded from clients. 

To understand the performance of FedReID, we compare global and local models with the other two models: 1) \textit{Standalone training}: the models trained in clients with their own dataset (without participating in federated learning); 2) \textit{Centralized training}: the model trained using the combination of all benchmark datasets, simulating conventional person ReID training that centralizes datasets. Centralized training can be treated as the upper bound of FedReID. While FedReID is meaningful for a client only when the performance of global or local models is better than standalone training.  


\subsubsection{Impact of Federated Settings} We first investigate the performance of FedReID (the global model) using the FedPav algorithm under different federated settings, including batch size $B$ and local epochs $E$. 

\begin{figure}[t]
  \centering
  \begin{subfigure}[b]{0.49\linewidth}
    \includegraphics[width=\linewidth]{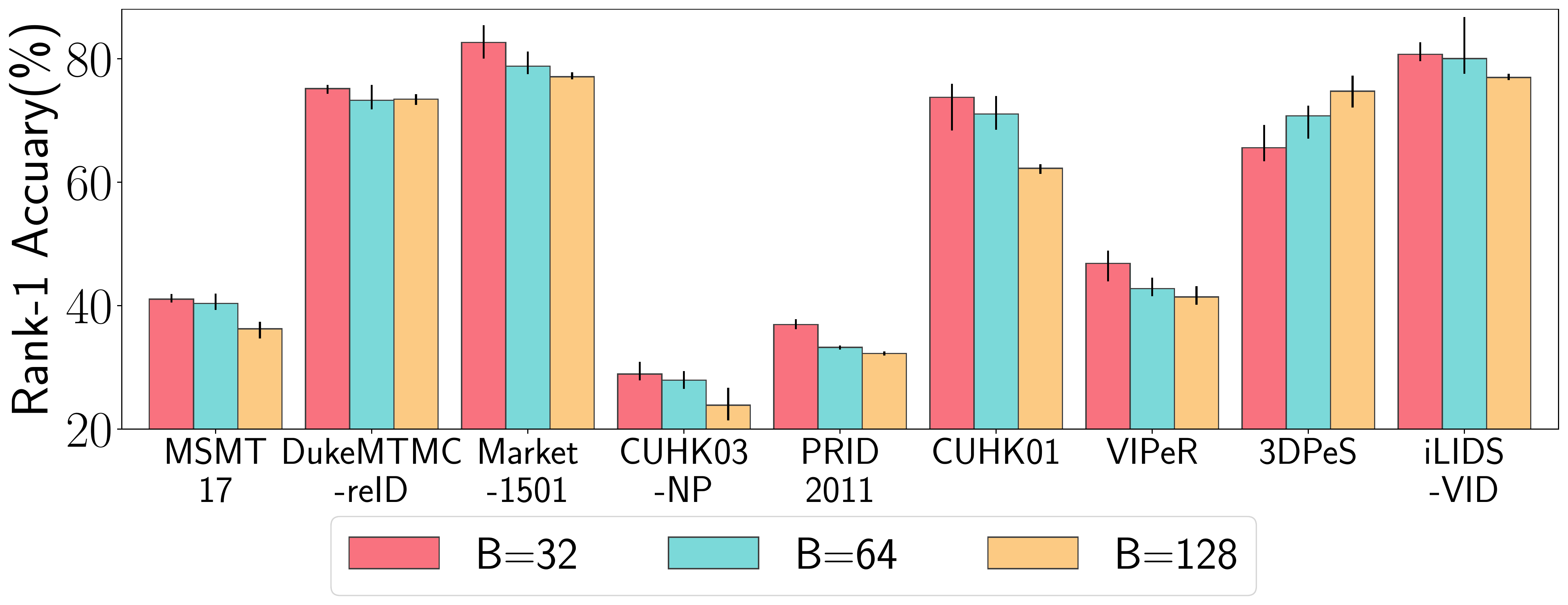}
    \caption{Batch Size $B$}
    \label{fig:batch-size}
  \end{subfigure}
  \hfill
  \begin{subfigure}[b]{0.49\linewidth}
    \includegraphics[width=\linewidth]{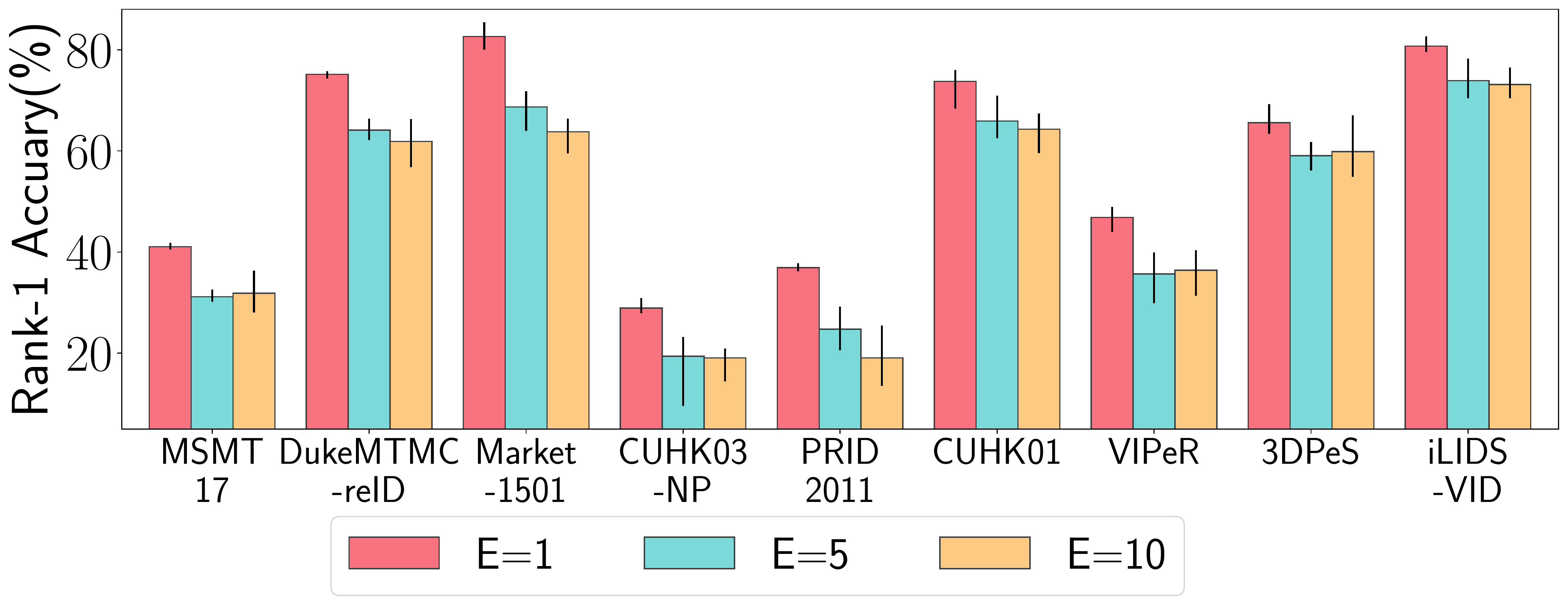}
    \caption{Local Epoch $E$}
    \label{fig:local-epoch}
  \end{subfigure}
  \caption{Performance comparison of (a) different batch sizes and (b) different local epochs. Batch size $B = 32$ and local epoch $E = 1$ achieves better overall performance than other settings. We run a total of 300 epochs for these experiments.}
  \label{fig:federated-settings}
\end{figure}

Batch size reflects the trade-off between computation power consumption and model accuracy. With the same local epochs, a larger batch size reduces computation time because the training can better take advantage of the parallelism provided by the client hardware. (Computation is fully utilized as long as $B$ is large enough.) Figure \ref{fig:batch-size} compares the rank-1 accuracy of FedPav using different batch sizes $B = \{32, 64, 128\}$, under the setting that local epochs $E = 1$ and communication rounds $T = 300$. Smaller batch size generally achieves better performance in most datasets, while consumes higher computation.


Local epochs reflect the trade-off between the communication cost and model accuracy. The total training epochs $E_{total}$ can be calculated with $E_{total} = T \times E$, where T is the communication rounds and $E$ is the number of local epochs. By fixing the total training epochs for a fair comparison, smaller $E$ means larger communication rounds $T$, requiring higher communication costs. Besides, we compare the rank-1 accuracy of different numbers of local epochs in Figure \ref{fig:local-epoch}. Despite that $E = 5$ performs worse than $E = 10$ in several datasets, smaller numbers of local epochs $E$ generally result in better performance. The smallest number of local epoch $E = 1$ achieves much better performance than $E = 5$ and $E = 10$ in all datasets, while it requires the highest communication cost, indicating the trade-off between communication costs and model accuracy. 



\subsubsection{Impact of Statistical Heterogeneity} 
\label{sec:impact-stats}
The statistical heterogeneity hinders the convergence and performance of FedReID. Specifically, non-IID causes difficulty in convergence and both non-IID and unbalanced data limits the performance of FedReID. 

Figure \ref{fig:convergence} shows that FedPav does not converge well as the accuracy (of the global model) fluctuates throughout training. We argue that it is mainly due to non-IID data of nine clients. As datasets in clients have domain discrepancy (e.g. illumination, resolution, scenes, etc.), aggregating them simply by weighted average leads to unstable and unpredictable results. As a result, it causes difficulty in selecting a representative global model for other scenarios. We report the accuracy by averaging the three best global models throughout training, evaluated every ten rounds.

Furthermore, Table \ref{tab:fedpav} compares the performance of the global and local models obtained from FedReID with standalone and centralized training. The results are two-fold: On the one hand, standalone training outperforms both the global and local models in large datasets such as DukeMTMC-reID \cite{zheng2017dukemtmc-reid} and CUHK03-NP \cite{Li2014CUHK03}; On the other hand, both the global and local models outperform standalone training in small datasets such as VIPeR \cite{Gray2008ViewpointIP} and 3DPeS \cite{3dpes}, and even outperforming centralized training in iLIDS-VID dataset \cite{iLIDS-VID}. These results indicate that although clients with larger datasets do not benefit from FedReID, the ones with smaller datasets gain significant improvement. We interpret the results from two perspectives: 1) For clients with large datasets, they dominate in server aggregation as the weights for aggregation are positively correlated with data volumes, causing less gain from others; 2) For clients with small datasets, they learn from other clients more effectively because their models are not well trained.

Another observation from Table \ref{tab:fedpav} is that local models outperform the global model in all datasets. As the global model is produced by aggregating local models, we argue that non-IID data causes performance degradation in the server aggregation. Better aggregation methods can be considered to better transfer knowledge from local models to the global model.




\begingroup
\setlength{\tabcolsep}{0.3em}
\begin{table}[t]
\caption{Rank-1 accuracy comparison of the global model and local models obtained from FedReID, standalone training, and centralized training. FedReID effectively improves the performance on small datasets. However, it performs worse than standalone training on large datasets due to statistical heterogeneity. We run the experiment with $B = 32$ and $E = 1$.}
\begin{center}
\begin{tabular}{lccccccccc}
\toprule    
Methods & MSMT17 & DukeMTMC & Market & CUHK03 & PRID2011 & CUHK01 & VIPeR & 3DPeS & iLIDS-VID
\\
\midrule
Centralized Training & 54.6 & 84.2 & 91.7 & 64.0 & 80.0 & 89.7 & 65.5 & 82.1 & 80.6 \\
\midrule
Standalone Training & \textbf{49.6} & \textbf{80.1} & \textbf{88.9} & \textbf{49.3} & \textbf{55.0} & 69.0 & 27.5 & 65.4 & 52.0 \\
Global Model & 41.0 & 74.3 & 83.4 & 31.7 & 37.7 & 73.4 & 48.1 & 69.2 & 79.9 \\
Local Model & 48.3 & 78.1 & 83.6 & 39.5 & 50.7 & \textbf{80.7} & \textbf{52.0} & \textbf{80.6} & \textbf{84.7} \\
\bottomrule
\end{tabular}
\end{center}
\label{tab:fedpav}
\end{table}
\endgroup

\section{Performance Optimization}
\label{sec:performance-optimization}

In this section, we first propose three methods to address the problems caused by statistical heterogeneity: client clustering, knowledge distillation, and dynamic weight adjustment. Then, we present experimental results of these optimization methods, compared with standalone training and the benchmark results. 

\subsection{Client Clustering}
\label{sec:clustering}

To tackle the performance degradation caused by non-IID data in server aggregation of all clients, we propose to aggregate clients with similar data distributions. As discussed in Section \ref{sec:impact-stats}, local models outperform the global model in all datasets. The global model is obtained by aggregating local models, so the performance drop mainly sources from the aggregation of clients with diverse data distributions. To tackle this problem, we propose \textit{client clustering} to split clients into several groups based on their data distributions and aggregate models within each group in the server.



\begin{figure}[t]
  \centering
  \begin{subfigure}[b]{0.31\linewidth}
    \includegraphics[width=\linewidth]{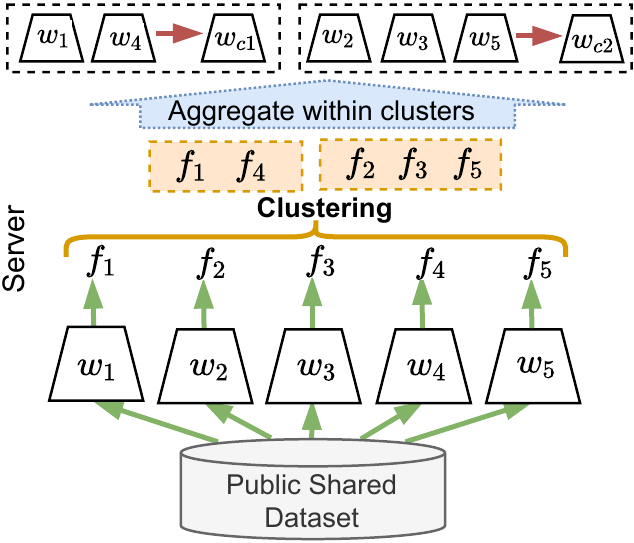}
    \caption{Client Clustering (CC)}
    \label{fig:clustering}
  \end{subfigure}
  \hfill
  \begin{subfigure}[b]{0.31\linewidth}
    \includegraphics[width=\linewidth]{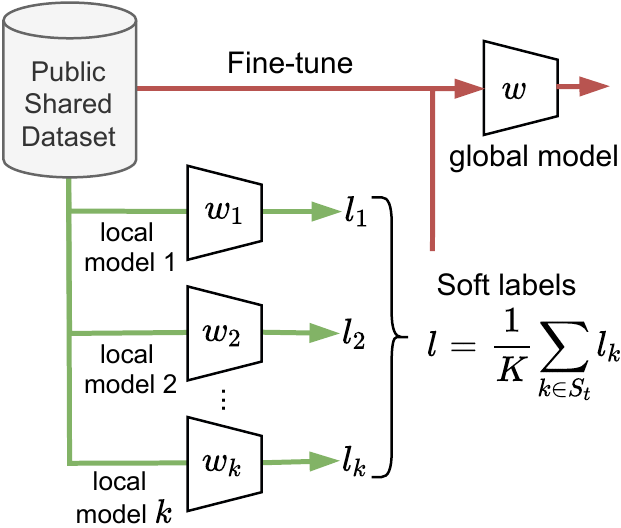}
    \caption{Knowledge Distillation (KD)}
    \label{fig:kd}
  \end{subfigure}
  \hfill
  \begin{subfigure}[b]{0.31\linewidth}
    \includegraphics[width=\linewidth]{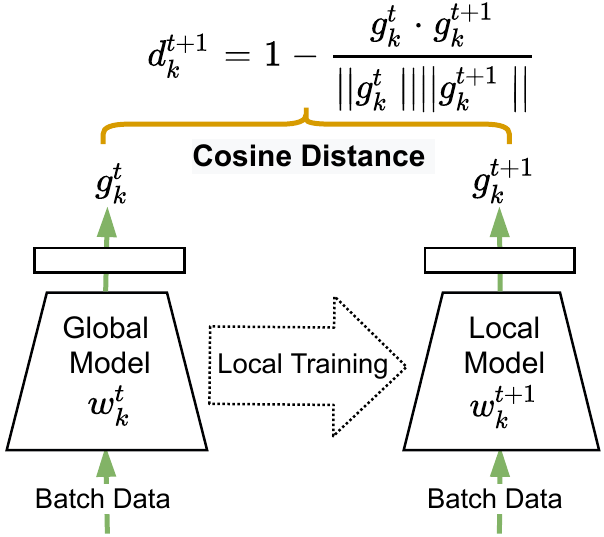}
    \caption{Cosine Distance Weight (CDW)}
    \label{fig:cdw}
  \end{subfigure}
  \hfill
  \caption{Illustrations of proposed performance optimization methods: (a) client clustering, (b) knowledge distillation, and (c) cosine distance weight.}
  \label{fig:optimization}
\end{figure}

Figure \ref{fig:clustering} depicts the process of client clustering with the following steps: (1) We extract features $f_k$ from one batch data (32 samples) of a public person ReID dataset \footnote{The public person ReID dataset is shareable among the server and clients. This dataset can be unlabeled.} using the trained model $w_k$ from client $k$. (2) We adopt a clustering algorithm to cluster these features into multiple groups. (3) We aggregate models of clients within each group, obtaining a global model in each group. (4) We use the global model of each group to update local models of clients within that group for the next training round. In Figure \ref{fig:clustering}, we cluster clients into two groups: one group contains clients \{1, 4\} and another one contains clients \{2, 3, 5\}, based on their features $f$. Then, we aggregate $w_1$ and $w_4$ to obtain global model $w_{c1}$, and aggregate $w_2$, $w_3$, and $w_5$ to obtain $w_{c2}$. At the start of next training round, we update local models of clients \{1, 4\} with $w_{c1}$ and local models of clients \{2, 3, 5\} with $w_{c2}$. Client clustering obtains multiple global models after training, so we focus on evaluating personalized local models $w_k$ of each client $k$.



In this way, we use the features as a proxy to measure the similarity of data distributions among clients. The intuition behind client clustering is that the clients clustered into the same group share more similar data distributions. The choice of the clustering algorithm is important for the overall performance of this method. 
We utilize a hierarchical clustering algorithm, FINCH \cite{sarfraz2019finch}, to cluster clients based on similarities of features extracted from their models. Regarding each client as a cluster at the start, we group the clients that are first neighbors; 
Two clients are first neighbors if their features have the shortest distance (cosine similarity) or they share the same first neighbor. FINCH merges first neighbors in each clustering step. In our scenario, since nine clients would be merged into one cluster after two to three clustering steps, we only cluster for one step per communication round. As a result, the server would have two to three clusters, where each cluster contains two to seven clients. FINCH is able to deliver good clustering results without prior knowledge of the targeted number of clusters.

\subsection{Knowledge Distillation}
\label{sec:kd}

Besides client clustering, we adopt \textit{knowledge distillation} (KD) to elevate performance and improve the convergence of FedReID. Since local models outperform the global model, it suggests that local models contain more knowledge than the global model --- simple server aggregation could not effectively aggregate knowledge from local models. Knowledge distillation is a method proposed by Hinton et, al. \cite{kd} to transfer knowledge from a teacher model to a student model, where the teacher model contains more knowledge than the student model. We adopt knowledge distillation to better transfer knowledge from local models to the global model, regarding clients as teachers and the server as the student. 

After clients finish local training and upload models, we apply knowledge distillation with a public shared dataset $\mathcal{D}_{shared}$ in the server. Figure \ref{fig:kd} illustrates the additional steps required from knowledge distillation: (1) The server uses each trained model $w_k$ of client $k$ to generate soft labels \footnote{These labels are termed soft labels as they are the predicted labels, not the actual labels, of the dataset.} $\ell_k$ using samples of $\mathcal{D}_{shared}$. These soft labels represent the knowledge of clients' models. (2) Apart from model aggregation, the server aggregates these soft labels with $\ell = \frac{1}{K} \sum_{k \in S_t} \ell_k$. (3) The server fine-tunes the global model with $\mathcal{D}_{shared}$ and corresponding labels $\ell$ to learn the distilled knowledge.


\subsection{Weight Adjustment}
\label{sec:weight-adjustment}


In addition to tackling the performance degradation caused by non-IID data, we propose to dynamically update the weights for aggregation to curb the adverse effect of unbalanced data. As discussed in Section \ref{sec:impact-stats}, the weights of server aggregation are inappropriate. The formula for server aggregation \cite{McMahanMRHA17,zhuang2020fedreid} is $w^{t+1} = \sum_{k \in S_t} \frac{n_k}{n} w^{t+1}_k$, where $n$ is the total data volume and $n_k$ is the data volume of client $k$. The weights of local models depend on the data volume of clients --- larger datasets lead to larger weights. Since data volumes have large discrepancies among datasets, large datasets dominate in the server aggregation. For example, the weight of the largest datasets (MSMT17 \cite{Wei2017Msmt} dataset) is around 40\%, while the weight of smallest dataset (iLIDS-VID \cite{iLIDS-VID}) is only 0.3\%. Models from smaller datasets are almost negligible in aggregation. Such unbalanced data volume hampers clients with large datasets to effectively learn from others. Hence, we introduce a novel weight adjustment method to obtain more suitable weights for weighted average in aggregation.


\textbf{Cosine Distance Weight} We introduce \textit{cosine distance weight} (CDW) to substitute the weight of data volumes. CDW adjust the weights for aggregation dynamically in each round, based on how well models are trained in clients. It is measured by changes in features extracted from models before and after training. Such changes are calculated by cosine distance. Particularly, in each training round, client $k$ downloads and trains on the global model $w^t_k$ from the server to obtain a new local model $w^{t+1}_k$. Figure \ref{fig:cdw} demonstrates our method to calculate the new weight with $w^t_k$ and $w^{t+1}_k$, with following steps: 1) Client $k$ extracts logits $g^t_k$ with a random batch data $\mathcal{D}_{batch}$ using $(w^t_k, v^t_k)$ \footnote{$(w^t_k, v^t_k)$ is the concatenation of global model $w^t_k$ and local classifier $v^t_k$}. 2) Client $k$ obtains new local model $(w^{t+1}_k, v^{t+1}_k)$ after local training. 3) Client $k$ extracts features $g^{t+1}_k$ with $\mathcal{D}_{batch}$ using  $(w^{t+1}_k, v^{t+1}_k)$. 4) We calculate the cosine distance of these two logits $g^{t}_k$ and $g^{t+1}_k$, with following formula:

\begin{equation}
		d^{t+1}_k = 1 - \frac{g^{t}_k \cdot g^{t+1}_k}{\Vert g^{t}_k  \Vert \Vert g^{t+1}_k  \Vert},
\end{equation}

where the cosine distance $d^{t+1}_k$ of each client $k$ is pushed to the server. The server uses the formula below to obtain the new weight: 

\begin{equation}
		p^{t+1}_k =  \frac{d^{t+1}_k}{\sum_{k \in S_t} d^{t+1}_k},
\end{equation}

where the server uses $p^{t+1}_k$ to replace $\frac{n_k}{n}$ as the new weight for aggregation. The intuition of CDW is that clients whose local trainings are more effective should contribute more to the aggregation. The cosine distance $d_k$ measures the scale of changes in local training that updates model $w^t_k$ to $w^{t+1}_k$.

\begin{figure}[t]
  \centering
  \begin{subfigure}[b]{0.24\linewidth}
    \includegraphics[width=\linewidth]{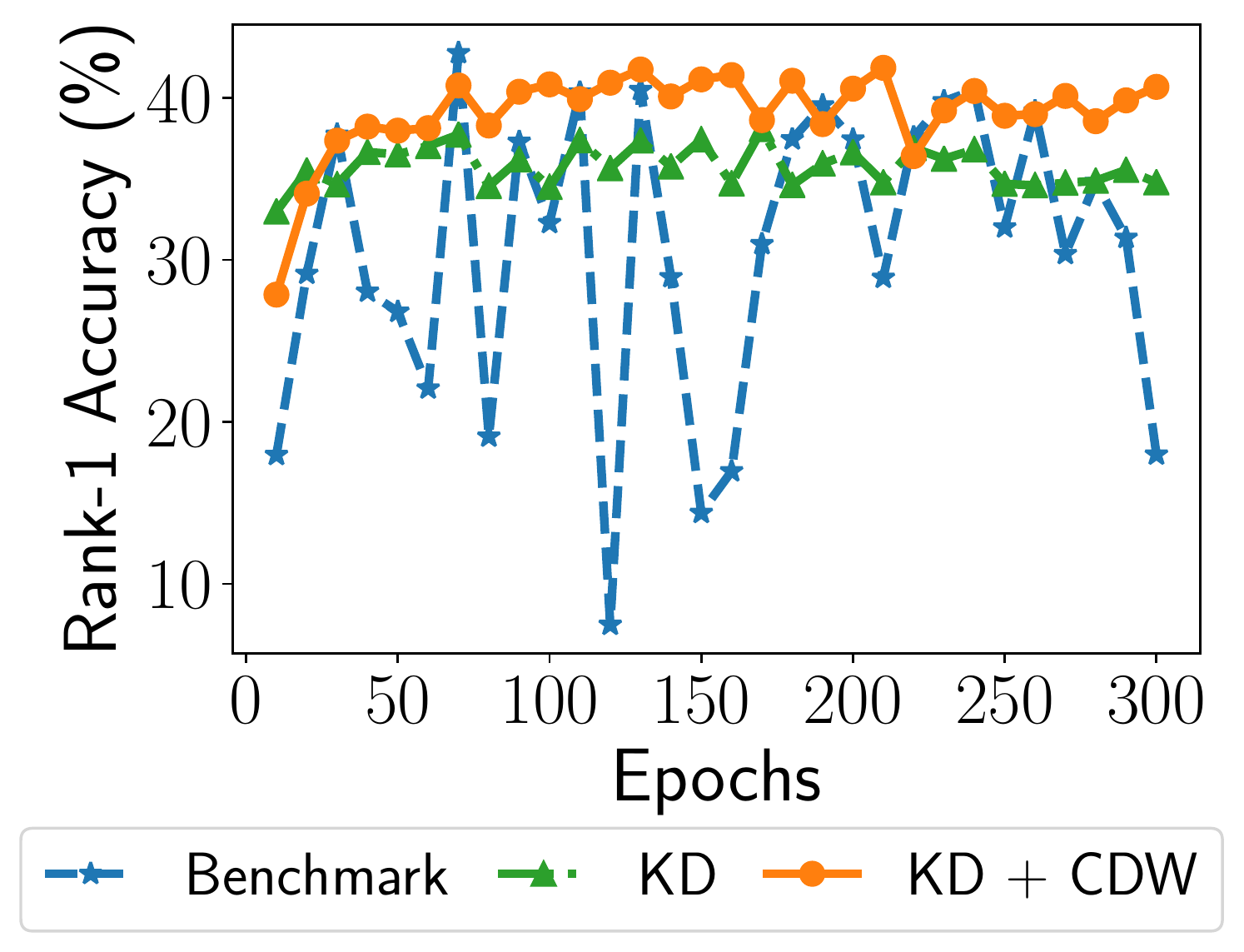}
    \caption{MSMT17}
  \end{subfigure}
  \hfill
  \begin{subfigure}[b]{0.24\linewidth}
    \includegraphics[width=\linewidth]{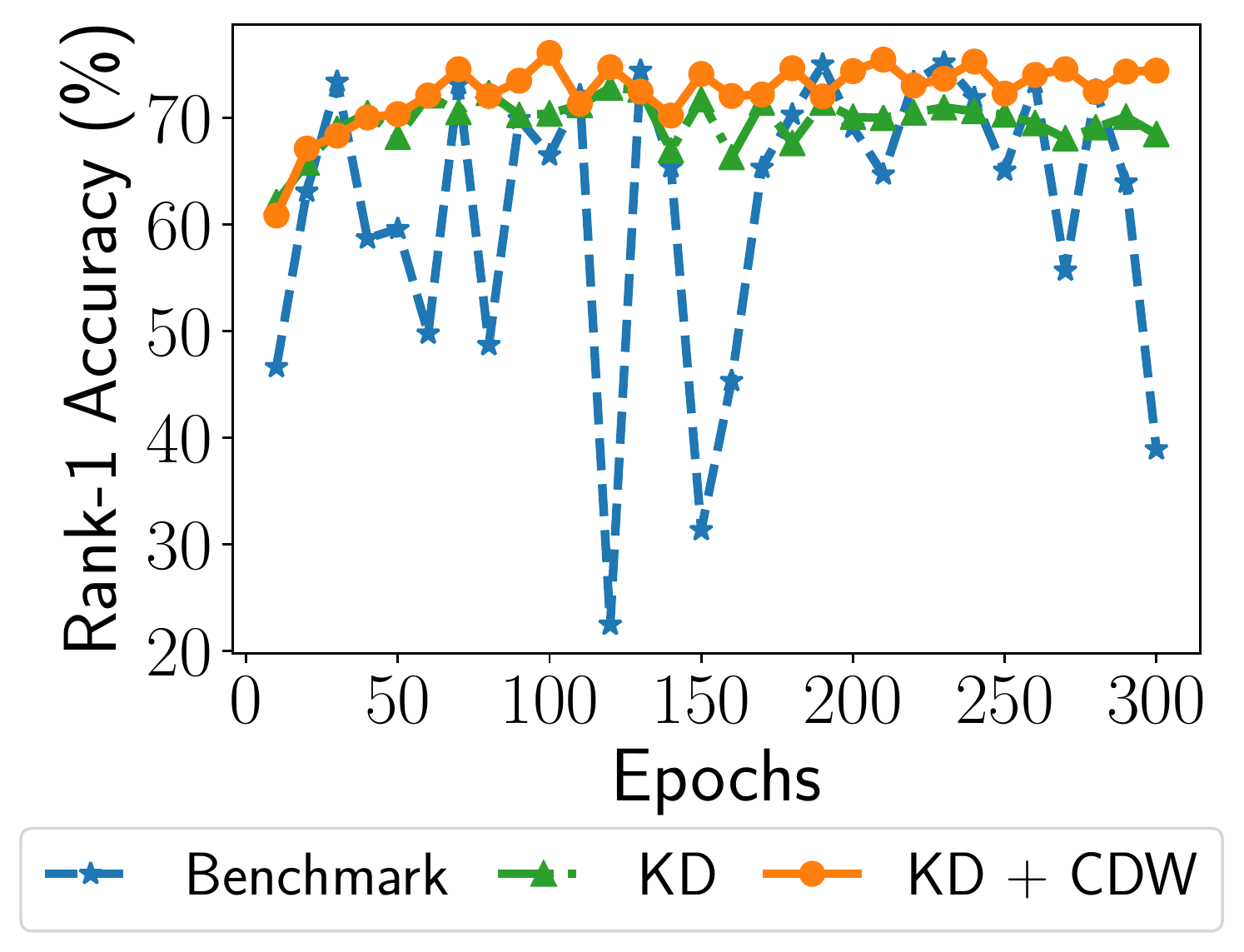}
    \caption{DukeMTMC-reID}
  \end{subfigure}
  \hfill
  \begin{subfigure}[b]{0.24\linewidth}
    \includegraphics[width=\linewidth]{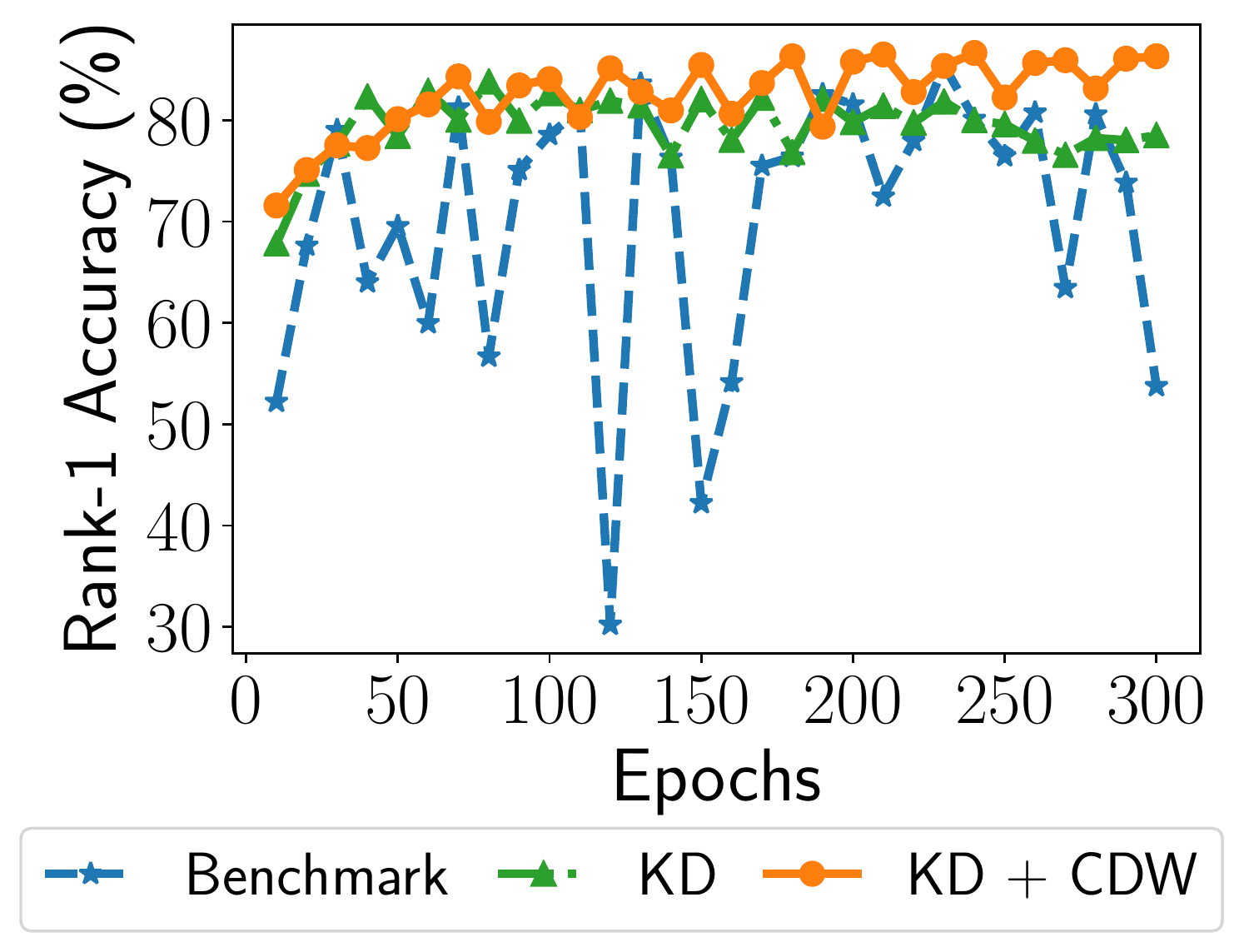}
    \caption{Market-1501}
  \end{subfigure}
  \hfill
  \begin{subfigure}[b]{0.24\linewidth}
    \includegraphics[width=\linewidth]{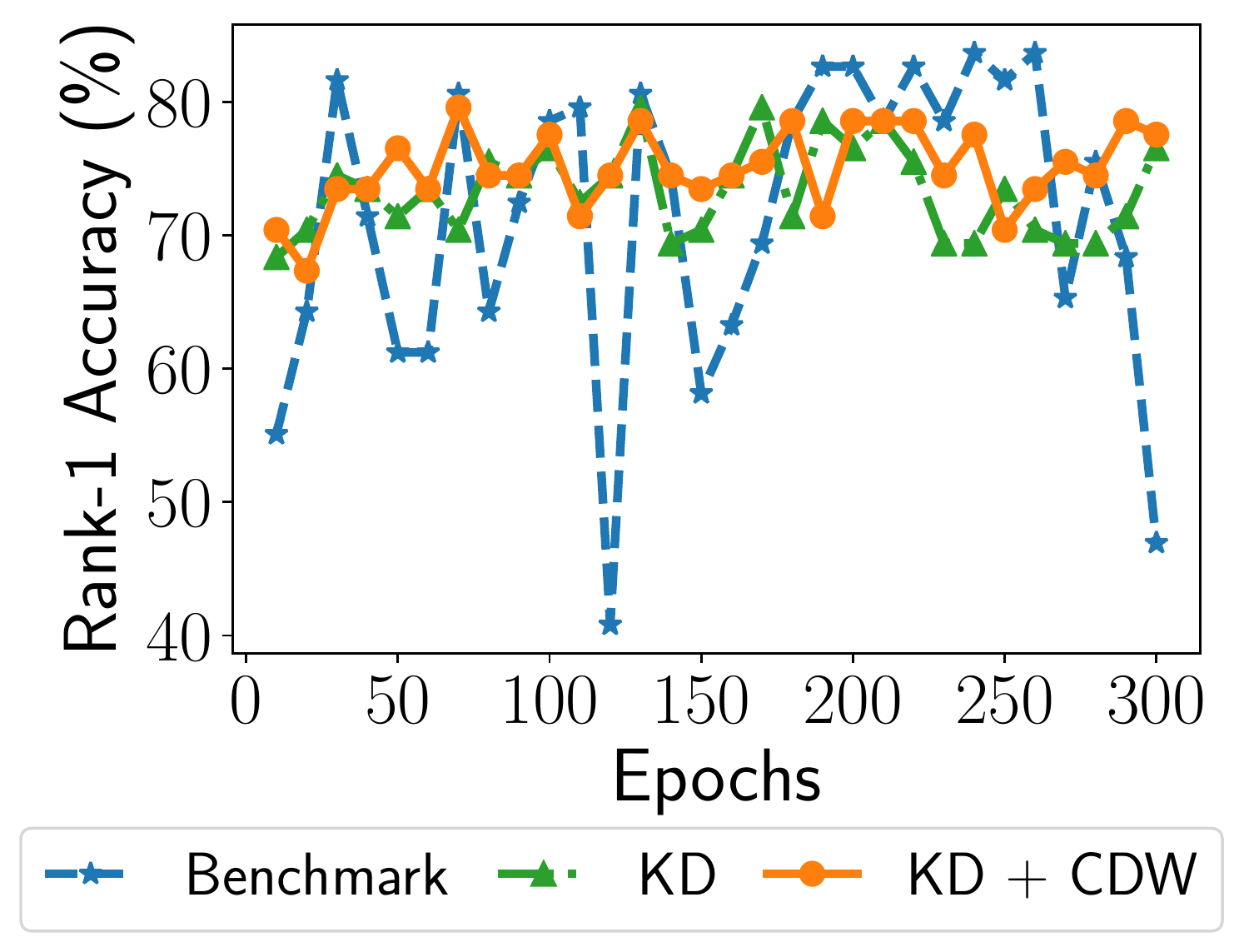}
    \caption{iLIDS-VID}
  \end{subfigure}
  \hfill
  \begin{subfigure}[b]{0.24\linewidth}
    \includegraphics[width=\linewidth]{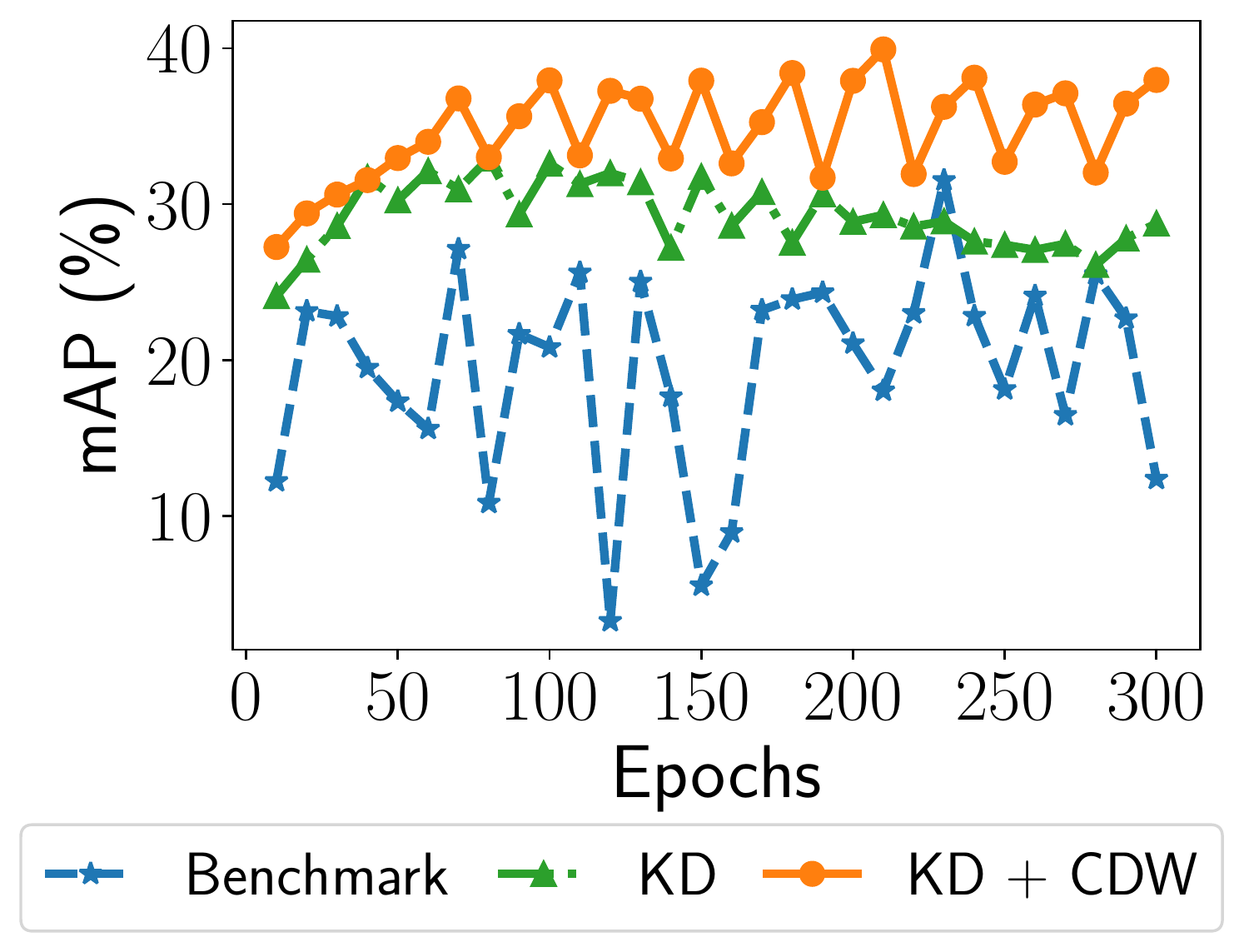}
    \caption{CUHK03-NP}
  \end{subfigure}
  \hfill
  \begin{subfigure}[b]{0.24\linewidth}
    \includegraphics[width=\linewidth]{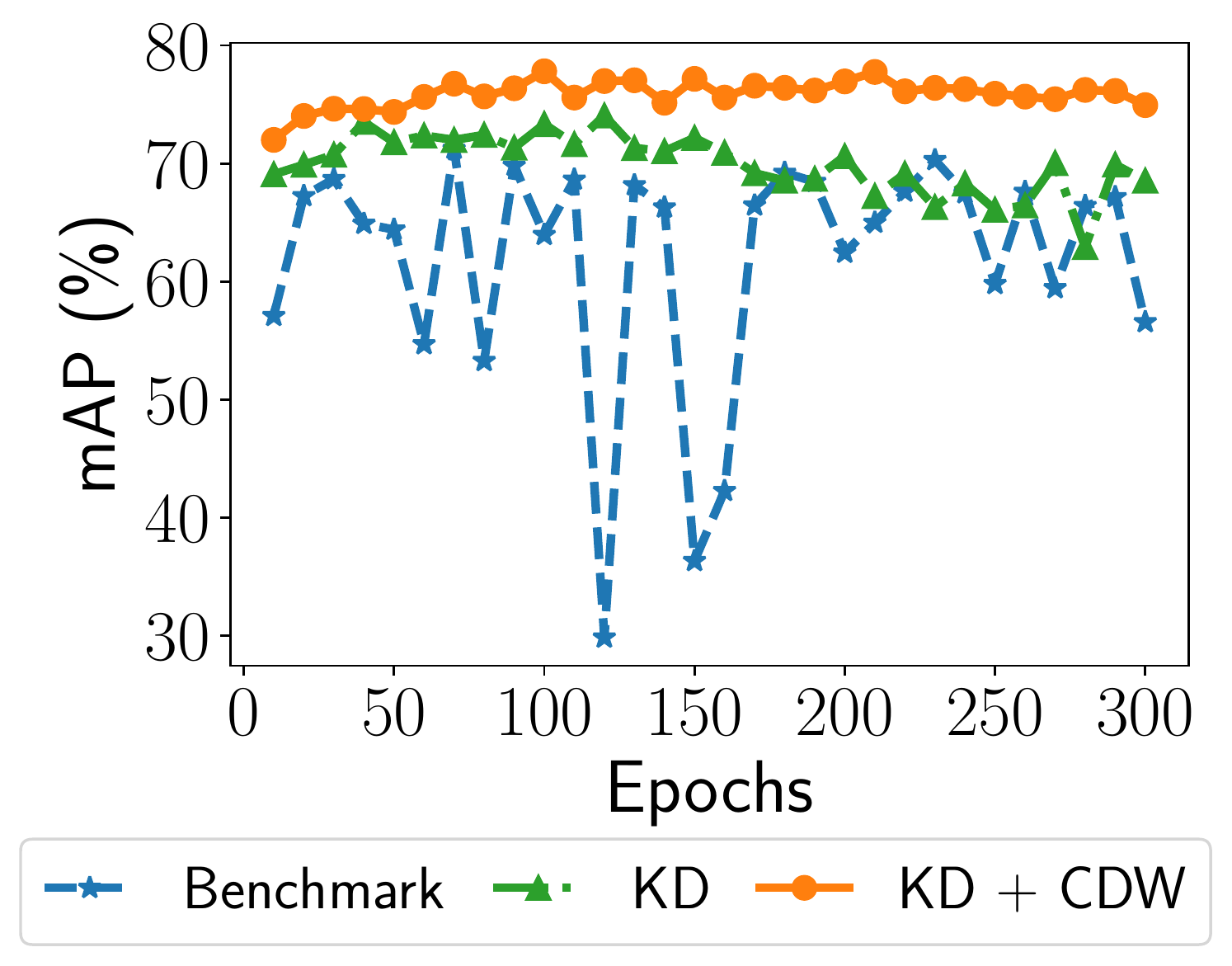}
    \caption{CUHK01}
  \end{subfigure}
  \hfill
  \begin{subfigure}[b]{0.24\linewidth}
    \includegraphics[width=\linewidth]{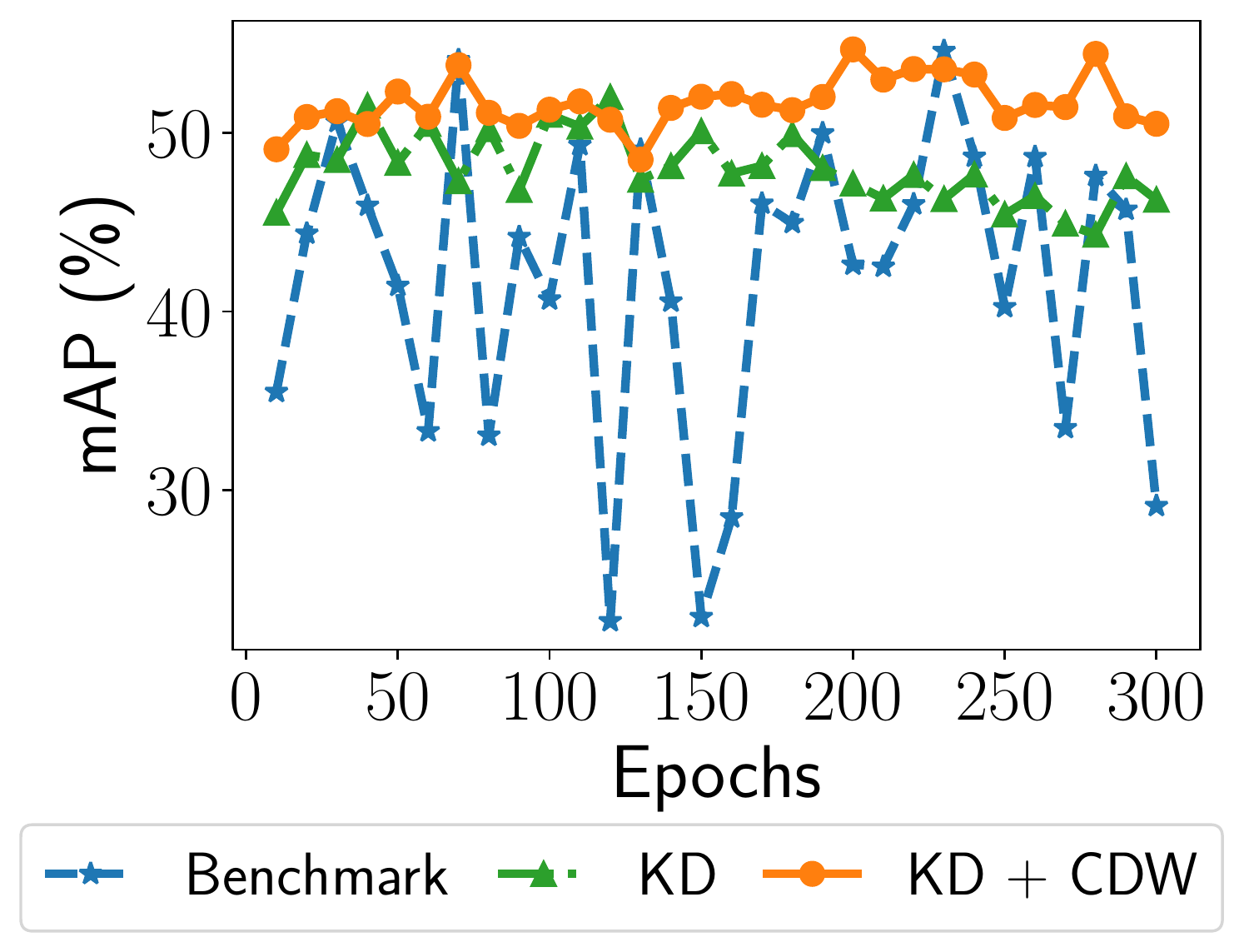}
    \caption{VIPeR}
  \end{subfigure}
  \hfill
  \begin{subfigure}[b]{0.24\linewidth}
    \includegraphics[width=\linewidth]{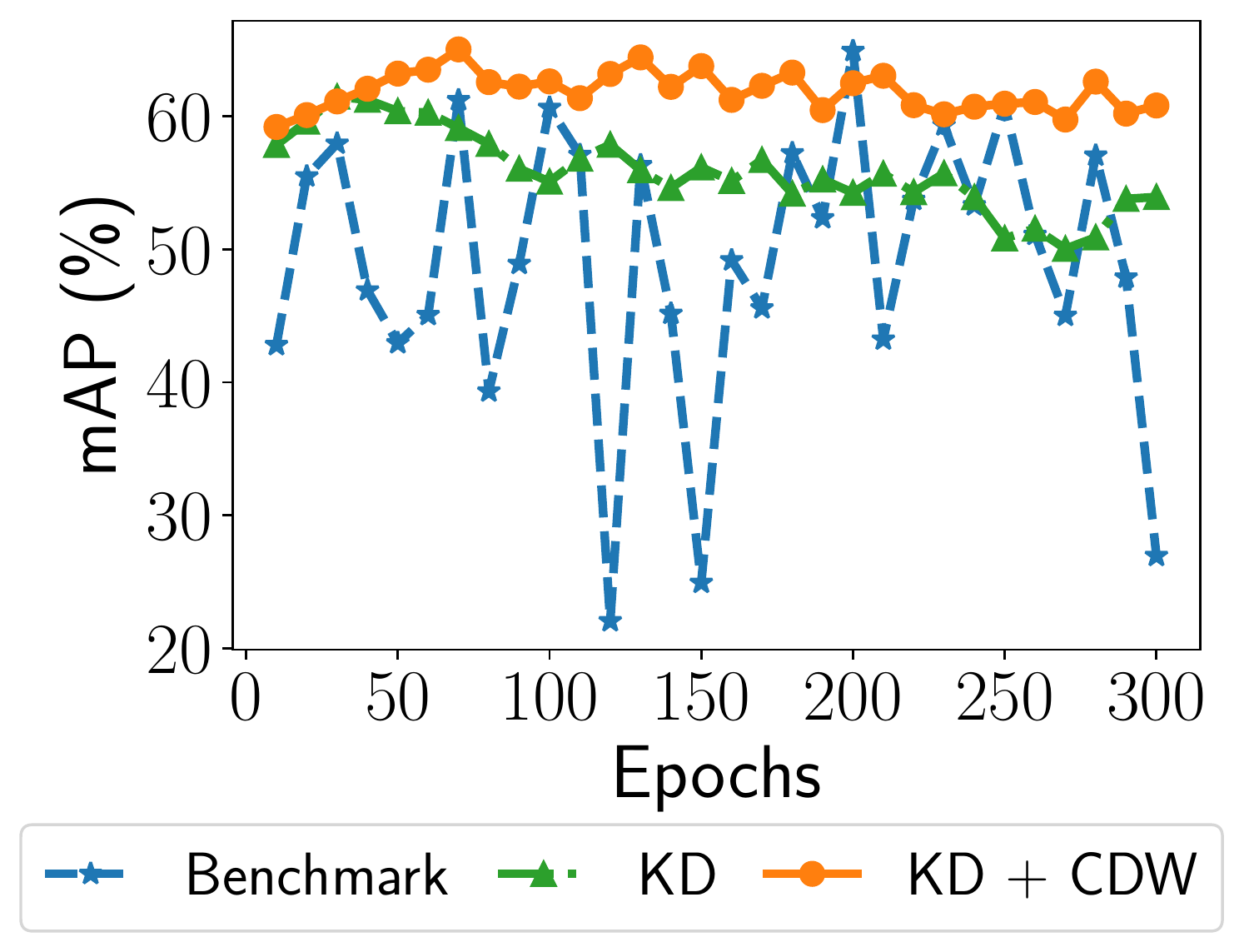}
    \caption{3DPeS}
  \end{subfigure}
  \hfill
  \caption{Convergence and performance (rank-1 accuracy and mAP) comparison of the benchmark results, knowledge distillation (KD), and combination of KD and cosine distance weight (CDW). KD effectively facilitates the convergence of FedReID. Besides, the combination of KD and CDW not only facilitates convergence but also effectively improves performance. These experiments are run with batch size $B = 32$ and local epoch $E = 1$.}
  \label{fig:convergence}
\end{figure}

\def\CDW{\leavevmode\rlap{\hbox to \hsize{\color{cyan!20}\leaders\hrule height 1\baselineskip depth 1.2ex\hfill}}}
\def\KD{\leavevmode\rlap{\hbox to \hsize{\color{pink!30}\leaders\hrule height 1\baselineskip depth 1ex\hfill}}}
\def\CC{\leavevmode\rlap{\hbox to \hsize{\color{yellow!20}\leaders\hrule height 1\baselineskip depth 1ex\hfill}}}

\begin{algorithm}[t]
  \caption{FedPav with \colorbox{yellow!20}{Client Clustering} and \colorbox{cyan!20}{Cosine Distance Weight}}
  \label{algo:fedpav-cc-cdw}
  \SetAlgoLined
  \KwInput{Total data volume $n$, client $k$'s data volume $n_k$, local epoch $E$, batch size $B$, shared dataset $\mathcal{D}_{shared}$}
  \KwOutput{Global model $w^T$, local models of each client $w^T_k$}
  
  \SetKwProg{Fn}{Server}{:}{}
  \Fn{\FnServer}{
      initialize $w^0$\;
      \For{each round t = 0 to T-1}{
      
          $S_t \leftarrow$ (random set of K clients selected from N clients) \Comment*[r]{Client selection}
          \For{each client $k \in S_t$ in parallel}{
              \CC $w_c^t \leftarrow$ (global model of the cluster client $k$ belongs to) \;
              $w^{t+1}_k , d_k \leftarrow $ \textbf{Client}($w_c^t$, $k$, $t$) \Comment*[r]{Distribution}
              \CC $f_k \leftarrow $ (extract features from $\mathcal{D}_{batch}$ with $w^{t+1}_k$) \;
          }
          
          \CC $C \leftarrow$ (cluster clients $S_t$ based on features $f_k$ for $k \in S_t$ with FINCH) \;
          \CC \For{each client $k$ in each cluster $c \in C$}{
              \CDW $d \leftarrow \sum_{k \in S_t} d_k$\;
              \CDW $w_c^{t+1} \leftarrow \sum_{k \in S_t} \frac{d_k}{d}
              w^{t+1}_k$ \Comment*[r]{Server aggregation}
          }
      }
      \KwRet best global model $w^{T}$, best client $k$'s local model $w^{T}_k$\;
  }
  
  \SetKwProg{Fn}{Client}{:}{}
  \Fn{\FnClient{$w$, k, t}}{
      $v \leftarrow$ (initialize classifier $v$ if $t == 0$, otherwise retrieve it from local store)\;
      $(w^{t}, v^{t}) \leftarrow (w, v)$ \Comment*[r]{Save a copy of global model before training}
      \For{each local epoch e = 0 to E-1}{
          \For{$b \in$ batches of local data in size $B$}{
              $(w, v) \leftarrow (w, v) - \eta \triangledown \mathcal{L}((w, v); b)$ \Comment*[r]{$(w, v)$ concatenation of two model structures}
          }
      }
      
      \CDW $g, g^t \leftarrow$ (extract logits with $(w, v)$ and $(w^{t}, v^{t})$ using $\mathcal{B}_{shared}$) \Comment*[r]{$\mathcal{B}_{shared}$ is a batch of $\mathcal{D}_{shared}$}
      
      \CDW $d = 1 - \frac{g^{t} \cdot g}{\Vert g^{t}  \Vert \Vert g  \Vert}$ \Comment*[r]{Cosine distance}
      
      

      store $v^{t}$\;
      \KwRet $w^{t}, d$ \Comment*[r]{Upload}
  }
\end{algorithm}

\begin{algorithm}[h]
    \caption{FedPav with \colorbox{pink!30}{Knowledge Distillation} and \colorbox{cyan!20}{Cosine Distance Weight}}
    \label{algo:fedpav-kd-cdw}
    \SetAlgoLined
    \KwInput{Total data volume $n$, client $k$'s data volume $n_k$, local epoch $E$, batch size $B$, shared dataset $\mathcal{D}_{shared}$}
    \KwOutput{Global model $w^T$, local models of each client $w^T_k$}
    
    \SetKwProg{Fn}{Server}{:}{}
    \Fn{\FnServer}{
        initialize $w^0$\;
        \For{each round t = 0 to T-1}{
            $S_t \leftarrow$ (random set of K clients selected from N clients) \Comment*[r]{Client selection}
            \For{each client $k \in S_t$ in parallel}{
                $w^{t+1}_k, d_k \leftarrow $ \textbf{Client}($w^t$, $k$, $t$) \Comment*[r]{Distribution}
                \KD $\ell_k \leftarrow$ (predict soft labels with $w^{t+1}_k$ and $\mathcal{D}_{shared}$)\;
            }
            
            \CDW $d \leftarrow \sum_{k \in S_t} d_k$\;
            \CDW $w^{t+1} \leftarrow \sum_{k \in C_t} \frac{d_k}{d} w^{t+1}_k$
            \Comment*[r]{Server aggregation}

            \KD $\ell \leftarrow \frac{1}{K} \sum_{k \in S_t} \ell_k$\;
            \KD $w^{t+1} \leftarrow $ (fine-tune $w^{t+1}$ with $\mathcal{D}_{shared}$ and soft labels $\ell$)\;
        }
        \KwRet best global model $w^{T}$, best client $k$'s local model $w^{T}_k$\;
    }
    
    \SetKwProg{Fn}{Client}{:}{}
    \Fn{\FnClient{$w$, k, t}}{
        $v \leftarrow$ (initialize classifier $v$ if $t == 0$, otherwise retrieve it from local store)\;
        $(w^{t}, v^{t}) \leftarrow (w, v)$\;
        \For{each local epoch e = 0 to E-1}{
            \For{$b \in$ batches of local data in size $B$}{
                $(w, v) \leftarrow (w, v) - \eta \triangledown \mathcal{L}((w, v); b)$ \Comment*[r]{$(w, v)$ concatenation of two model structures}
            }
        }
        
        \CDW $g, g^t \leftarrow$ (extract logits with $(w, v)$ and $(w^{t}, v^{t})$ using $\mathcal{B}_{shared}$) \Comment*[r]{$\mathcal{B}_{shared}$ is a batch of $\mathcal{D}_{shared}$}
        
        \CDW $d = 1 - \frac{g^{t} \cdot g}{\Vert g^{t}  \Vert \Vert g  \Vert}$ \Comment*[r]{Cosine distance}
        

        

        store $v$\;
        \KwRet $w, d$ \Comment*[r]{Upload}
    }
\end{algorithm}

\subsection{Combinations of Optimization Method}

We can achieve even better performance by combinations of these three optimization methods: client clustering (CC), knowledge distillation (KD), and cosine distance weight (CDW). We consider only combining CDW and CC, and CDW and KD. 

It is not desirable to combine client clustering and knowledge distillation because they both enhance the server aggregation. On the one hand, knowledge distillation only fine-tunes a single global model, while client clustering contains multiple global models. On the other hand, both methods address the non-IID problem: knowledge distillation aims to further improve the global model, while client clustering tends to elevate the performance of local models. Hence, we do not consider the combination of these two methods.

Since CDW tackles unbalanced data volume, either the combination of CDW and CC or the combination of CDW and KD addresses statistical heterogeneity with non-IID and unbalanced data problems. To combine CDW with CC or KD, we just need to replace the original weights in the server aggregation process with the new weights. As client clustering has no single global model, combining it with CDW aims to achieve better local models; As knowledge distillation further fine-tunes the global model, combining it with CDW aims to achieve a better global model. We summarize these two combinations in Algorithm \ref{algo:fedpav-cc-cdw} and Algorithm \ref{algo:fedpav-kd-cdw}.

\subsection{Evaluation}

We present the empirical evaluation of these performance optimization approaches, compared with the benchmark and standalone training. By default, we conduct these experiments with batch size $B = 32$ and local epoch $E = 1$. For both client clustering and knowledge distillation, we adopt an additional unlabelled dataset, CUHK02 \cite{cuhk02} dataset. This dataset is regarded as a public dataset that is shareable among clients and the server. CUHK02 dataset is an extension of the CUHK01 dataset. It includes 7,264 images of 1,816 identities collected from 6 camera views.

We first evaluate the effectiveness of knowledge distillation (KD) and the combination of CDW and KD by monitoring performance changes of global models as training proceeds. Figure \ref{fig:convergence} shows the performance changes (either rank-1 accuracy or mAP) of KD, the combination of CDW and KD, and the benchmark results on eight datasets. Compared with the benchmark results, training with KD achieves much better convergence; KD can also lead to higher performance, especially when datasets in clients share similar data distributions with the public shared dataset. For example, we use the CUHK02 dataset as the shared dataset, so the accuracy of the global models on both CUHK03-NP and CUHK01 datasets are better than the benchmark results. Moreover, training with the combination of KD and CDW achieves outstanding performance on almost all datasets --- better than the benchmark results or training with KD. These results indicate that the combination of KD and CDW is able to obtain the best generalized global model that is transferable to other scenarios.

\begingroup
\setlength{\tabcolsep}{0.3em}
\begin{table}[t]
\caption{Increase in rank-1 accuracy of benchmark results, client clustering (CC), cosine distance weight (CDW), and combination of CC and CDW, when comparing with standalone training. CC effectively improves the performance on larger datasets and CDW effectively elevates the performance on all datasets. Besides, the combination of CC and CDW achieves the best overall performance, especially on the larger datasets. These experiments are run with batch size $B = 32$ and local epoch $E = 1$.}
\begin{center}
\begin{tabular}{lccccccccc}
\toprule    
Methods & MSMT17 & DukeMTMC & Market & CUHK03-NP & PRID2011 & CUHK01 & VIPeR & 3DPeS & iLIDS-VID
\\
\midrule
Benchmark & -1.3 & -2.0 & -5.4 & -9.8 & -4.3 & +11.6 & +24.5 & +15.2 & +32.7 \\
FedNova \cite{wang2020fednova} & -2.1 & -2.8 & -4.4 & -14.6 & 0.0 & +9.9 & +24.4 & +12.6 & \textbf{+35.8} \\
FedProx \cite{fedprox} & -0.1 & -1.6 & +1.0 & -6.4 & -1.0 & +12.5 & +24.1 & +7.7 & +34.7 \\
CC & +2.4 & -1.3 & +0.1 & \textbf{+3.9} & +6.0 & +9.3 & +4.1 & -1.2 & +16.3 \\
CDW & +4.0 & +1.3 & +1.4 & +1.2 & +7.3 & \textbf{+13.8} & +26.0 & \textbf{+16.3} & +30.3 \\
CC \& CDW & \textbf{+4.1} & \textbf{+3.8} & \textbf{+2.0} & +2.2 & \textbf{+13.0} & +6.1 & \textbf{+28.2} & +6.5 & +28.6 \\
\bottomrule
\end{tabular}
\end{center}
\label{tab:local-model}
\end{table}
\endgroup

\begingroup
\setlength{\tabcolsep}{0.3em}
\begin{table}[t]
\caption{Rank-1 accuracy comparison of our proposed methods (CC \& CDW and KD \& CDW) with the existing approaches on two new datasets: CAVIAR \cite{caviar4reid} and GRID \cite{grid}. Our trained models outperform the existing methods on both datasets without extra fine-tuning. These results demonstrate the generalization ability of our methods.}
\begin{center}
  \begin{tabular}{lcccccccccccc}
  \toprule 
  \multicolumn{1}{l}{\multirow{2}{*}{Datasets}} &
  \multicolumn{8}{c}{Existing methods (w/o privacy except \cite{wu2021fedreid})} &
  \multicolumn{2}{c}{Ours (w/ privacy)} \\ \cline{2-8}  \cline{10-11}  
  \multicolumn{1}{c}{} & DSTML & UMDL & CrossGrad & MLDG & SSDAL & DIMN & Decentralized \cite{wu2021fedreid} & & CC \& CDW & KD \& CDW \\ 
  \midrule
  CAVIAR \cite{caviar4reid} & 28.2 & 41.6 & - & - & - & - & 45.6 & & 46.8 & 53.2 \\
  \midrule
  GRID \cite{grid} & - & - & 9.0 & 15.8 & 22.4 & 29.3 & 24.2 & & 30.0 & 36.8 \\
  \bottomrule
  \end{tabular}
  \end{center}
  \label{tab:generalization}
  \end{table}
  \endgroup

Next, we evaluate the effectiveness of client clustering (CC), CDW, and the combination of these two methods by comparing the performances of their local models. Table \ref{tab:local-model} shows the increase in rank-1 accuracy of several methods when compared with standalone training on nine datasets. Although FedNova \cite{wang2020fednova} and FedProx \cite{fedprox} slightly improve the performance of the smallest dataset (iLIDS-VID), they are incapable to elevate the performance of large datasets, like our benchmark method. We further analyze the results in three folds. Firstly, CC effectively mitigates the drawback of the benchmark, improving the performance on larger datasets such as MSMT17 \cite{Wei2017Msmt}. This is because the dominance of larger datasets over smaller datasets is reduced as they are clustered into different groups in aggregation. Most of the time, CC creates two clusters: the first one contains clients with PRID2011, CUHK03-NP, VIPeR, 3DPeS, and iLIDS-VID datasets; the second one contains clients with MSMT17, DukeMTMC-reID, Market-1501, and CUHK01 datasets. Secondly, CDW outperforms the standalone training on all datasets. It indicates that CDW effectively addresses the unbalanced data problem, such that all clients are beneficial in federated learning. Thirdly, the combination of CDW and CC further elevates the performance in most datasets. Although such combination produces slight decreases on smaller datasets compared with CDW, it significantly further improves the performance of larger datasets. It increases the motivation of clients with larger datasets to participate in federated learning.

Lastly, we demonstrate the generalization ability of our methods by comparing existing methods on CAVIAR4REID \cite{caviar4reid} and GRID \cite{grid} datasets. Specifically, we compare with unsupervised cross-domain fine-tuning methods: DSTML \cite{hu2015dstml} and UMDL \cite{peng2016umdl}, unsupervised generalization methods: CrossGrad \cite{shankar2018CrossGrad}, MLDS \cite{li2018mldg}, SSDAL \cite{su2016ssdal}, and DIMN \cite{song2019dimn}, as well as the recent work \cite{wu2021fedreid}. For evaluation on the CAVIAR4REID, we follow \cite{liu2014semi, peng2016umdl} to randomly select 36 identities that appeared on two camera views. GRID dataset contains 250 identities from two camera views. For both datasets, we use images of one camera view as the query and another one as the gallery. Table \ref{tab:generalization} shows that our proposed FedReID with optimizations (CC \& CDW and KD \& CDW) outperforms all existing methods on rank-1 accuracy on both datasets; KD \& CDW achieves especially good performance. Note that we do not fine-tune trained models on these two evaluation datasets. These results further illustrate the significance of our methods.

\section{Conclusion}
\label{sec:conclusion}

In this paper, we present Federated Person Re-identification (FedReID), a new paradigm of person ReID training with decentralized data. To investigate the challenges of FedReID, we construct a new benchmark to simulate real-world scenarios. Based on the results and insights from benchmark analysis, we propose three optimization approaches to elevate performance. We propose client clustering and knowledge distillation to address the non-IID problem and introduce cosine distance weight to address the unbalanced data problem. Empirical results demonstrate that the combination of cosine distance weight and client clustering achieves the best local models, and the combination of cosine distance weight and knowledge distillation achieves the best global model, among all methods. 
In the future, we plan to investigate the system heterogeneity challenges among clients. We also plan to extend FedReID to support unsupervised learning. 


\begin{acks}

This study is in part supported by the RIE2020 Industry Alignment Fund - Industry Collaboration Projects (IAF-ICP) Funding Initiative, as well as cash and in-kind contribution from the industry partner(s); Singapore MOE under its Tier 1 grant call, Reference number RG96/20; Natioanl Research Foundation, Prime Minister's Office, Singapore under its Sustainable Tropical Data Centre Testbed programme (STDCT), and its Central Gap Fund (“Central Gap” Award No. NRF2020NRF-CG001-027).
\end{acks}

\bibliographystyle{ACM-Reference-Format}
\bibliography{references}

\end{document}